\documentclass[runningheads,a4paper]{llncs}

\usepackage{amssymb}
\usepackage{graphicx}

\usepackage{amsmath}
\usepackage[hidelinks]{hyperref}
\usepackage{booktabs}
\usepackage{multirow}
\usepackage[numbers]{natbib}
\makeatletter 
\renewcommand\@biblabel[1]{#1.} 
\makeatother

\let\llncssubparagraph\subparagraph
\let\subparagraph\paragraph
\usepackage[compact]{titlesec}
\let\subparagraph\llncssubparagraph

\usepackage{mathptmx}
\usepackage[10pt]{moresize}

\usepackage{mathtools}

\usepackage{url}
\urldef{\mailsa}\path|{fady.medhat, david.chesmore, john.robinson}@york.ac.uk|    
\newcommand{\keywords}[1]{\par\addvspace\baselineskip
\noindent\keywordname\enspace\ignorespaces#1}

\begin{document}

\mainmatter  

\title{Music Genre Classification using \\ Masked Conditional Neural Networks}

\titlerunning{Music Genre Classification using Masked Conditional Neural Networks}

%
%
\author{Fady Medhat
\and David Chesmore\and John Robinson}
\authorrunning{Fady Medhat, David Chesmore, John Robinson}
\institute{Department of Electronic Engineering\\ University of York, York\\ United Kingdom\\
\mailsa}

%
%
\toctitle{Lecture Notes in Computer Science}
\tocauthor{Authors' Instructions}
\maketitle
\begin{abstract}
The ConditionaL Neural Networks (CLNN) and the Masked ConditionaL Neural Networks (MCLNN)\footnote[1]{Code: https://github.com/fadymedhat/MCLNN} exploit the nature of multi-dimensional temporal signals. The CLNN captures the conditional temporal influence between the frames in a window and the mask in the MCLNN enforces a systematic sparseness that follows a filterbank-like pattern over the network links. The mask induces the network to learn about time-frequency representations in bands, allowing the network to sustain frequency shifts. Additionally, the mask in the MCLNN automates the exploration of a range of feature combinations, usually done through an exhaustive manual search. We have evaluated the MCLNN performance using the Ballroom and Homburg datasets of music genres. MCLNN has achieved accuracies that are competitive to state-of-the-art handcrafted attempts in addition to models based on Convolutional Neural Networks.
\end{abstract}

\section{Introduction}
\vskip 0in
Automating the feature extraction is currently an active research field aiming to learn enhanced representations directly from the raw data rather than handcrafting them. Neural Network based architectures have been used in this regard for image recognition \citep{RN367} and sound \citep{RN44}. The adoption of these architectures to sound recognition usually occurs after they gain wide acceptance in other application domains such as image recognition. For example, stacked Restricted Boltzmann Machines (RBM) \citep{RN173} forming a Deep Belief Net (DBN)\citep{RN304} to extract features were initially introduced to showcase the capability of these stacked generative layers to be used as a dimensionality reduction technique when applied on images of handwritten digits. Later, Hamel et al. \citep{RN290} trained a DBN of three RBM layers over frames of a spectrogram to extract abstract representations from music files that were classified using a Support Vector Machine (SVM) \citep{RN237} for a music genre classification task. Convolutional Neural Networks (CNN) as well were initially introduced in the work of LeCun et al. \citep{RN307} for images, and later attempts followed to use it for sound \citep{RN340, RN316,RN375}

Despite the success of these architectures for images, they are not designed to exploit the time-frequency representation of sound efficiently. For example, DBNs ignore the inter-frames relation by treating a spectrogram's frame in isolation from neighboring frames, and CNNs depend on weight sharing, which does not preserve the spatial locality of the learned features.

The ConditionaL Neural Networks (CLNN) \citep{RN465} and the Masked ConditionaL Neural Networks (MCLNN) \citep{RN465} are designed to preserve the spatial locality of the learned features, where there is a dedicated link for every feature in a feature vector compared to the weight sharing using the CNN. The CLNN preserve the temporal relation between the frames by considering a window rather than the isolated frame used in the RBM, and the mask in the MCLNN enforces a systematic sparseness over the network's links. The mask design follows a band-like pattern, which allows the network to be frequency shift-invariant mimicking a filterbank. Additionally, the mask explores several feature combinations concurrently analogous to handcrafting the optimum combination of features through a mix-and-match operation, while preserving the spatial locality of the features.

\section{Related Models} 
\vskip -0.2in
\begin{figure}
\begin{minipage}{0.48\textwidth}
\vskip -0.3in
\begin{center}
\centerline{\includegraphics[width=6.0cm]{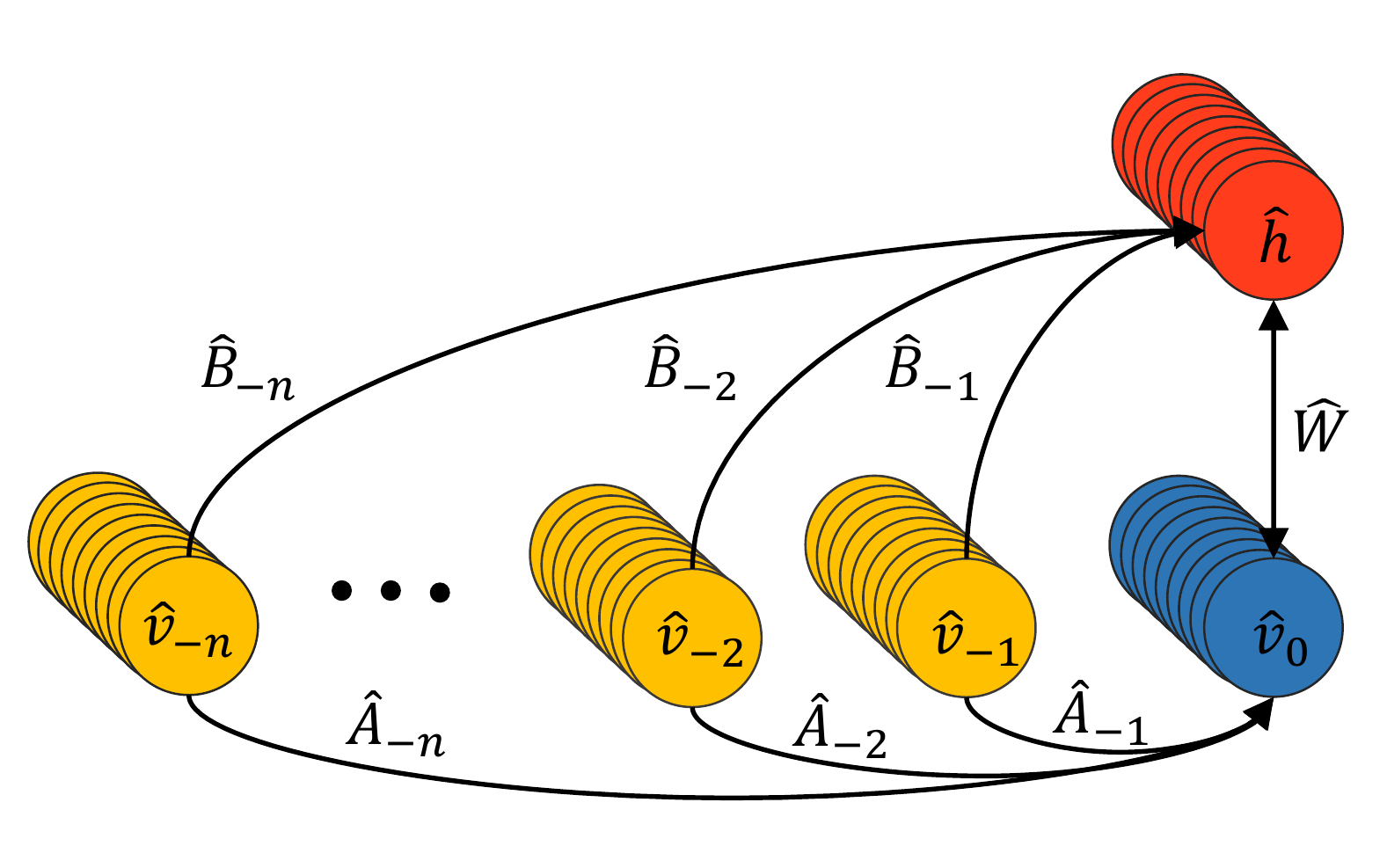}}
\vskip -0.2in
\caption{Conditional RBM}
\label{fig:crbm}
\end{center}
\end{minipage}\hfill
\begin{minipage}{0.48\textwidth}
\begin{center}
\centerline{\includegraphics[width=6.0cm]{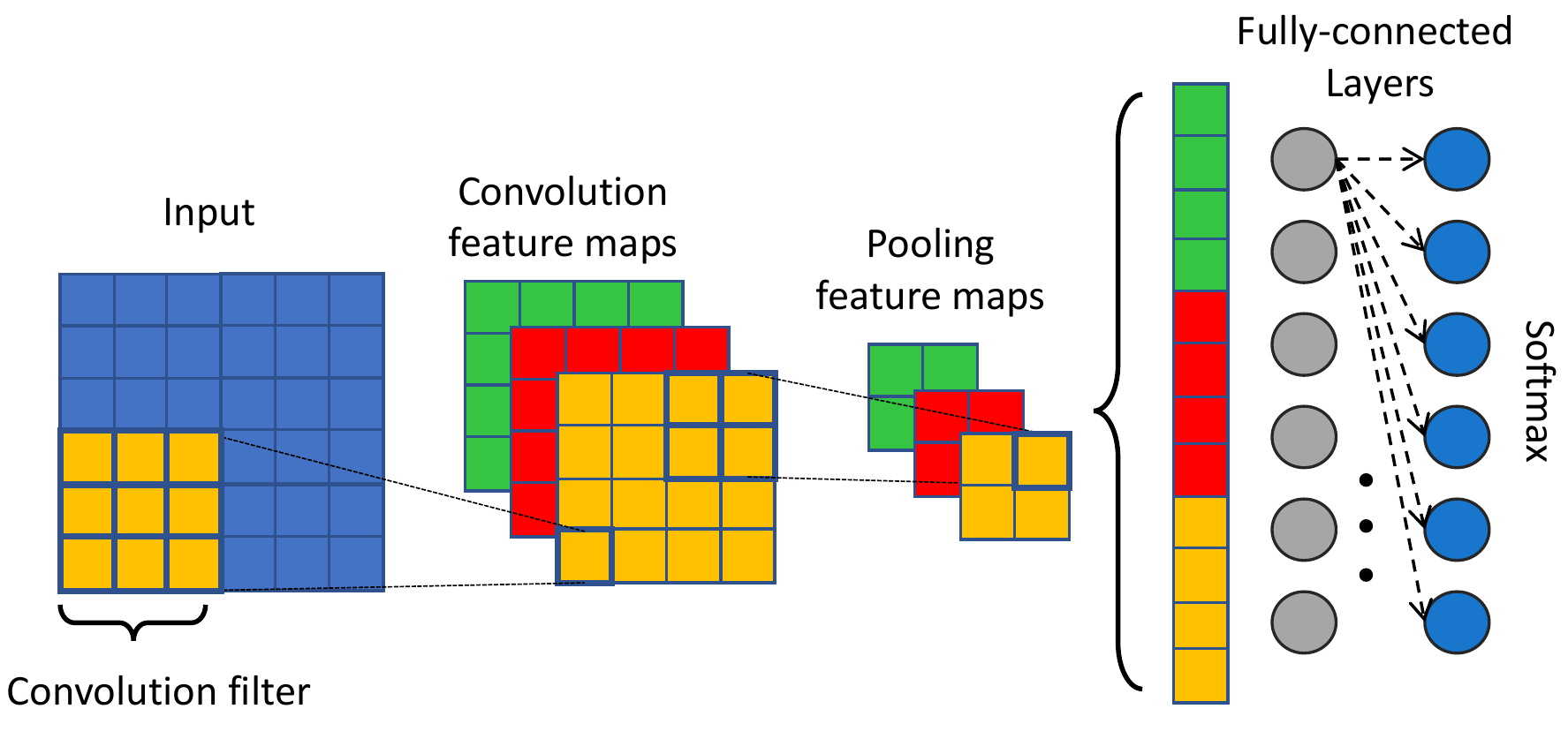}}
\vskip -0.2in
\caption{Convolutional Neural Network}
\label{fig:cnn}
\end{center}
\end{minipage}\hfill
\vskip -0.4in
\end{figure} 
The Conditional Restricted Boltzmann Machine (CRBM) \citep{RN207} by Taylor et al. extended the RBM to the temporal dimension to allow an RBM to learn about a temporal window of frames rather than being trained on static bag-of-frames. To fulfill this aim, the CRBM adapted conditional links to capture the influence of the previous frames on the current one. Fig. \ref{fig:crbm} shows a CRBM layer, where the normal RBM is represented with the bidirectional connections $\hat{W}$ going across the visible vector $\hat{v}_0$ and the hidden nodes $\hat{h}$. The $\hat{B}$ links in the figure represent the conditional links from the previous visible vectors ($\hat{v}_{-1},\hat{v}_{-2},...,\hat{v}_{-n} $) to the hidden layer $\hat{h}$. Similarly, the $\hat{A}$ links capture the autoregressive relation from the previous visible vectors to the current one $\hat{v}_0$. Layers of a CRBM can be stacked over each other similar to a DBN, where Taylor et al. trained a CRBM to model the human motion over a multichannel signal of human joints activity. Mohamed et al. \citep{RN257} extended the CRBM with the Interpolating Conditional Restricted Boltzmann Machine (ICRBM), which showed an enhanced performance by including the influence of the future frames in addition to the past ones for phoneme recognition. The work of Battenberg et al. \citep{RN209} was another attempt to use the CRBM for sound, where they used the CRBM to analyze drum patterns.

Similar modifications were introduced to the CNN to fit the time-frequency representation. The CNN architecture, shown in Fig. \ref{fig:cnn}, is based on the two primary operations: convolution and pooling. The convolution operation scans the 2-dimensional representation with a small weight matrix (or filter), e.g. $5\times5$, where a form of a weighted sum is generated from the element-wise multiplication between the filter and the region of the image being scanned. The output of each step of the filter is a scalar value positioned in a new representation of the image known as the feature map. The convolutional layer generates several feature maps. The number of feature maps matches the number of filters used. Mean or max pooling follows the convolution to reduce the resolution of the feature maps. These two operations are consecutively repeated to form a deep architecture of a CNN, where the output of the final layer is flattened to a single feature vector to be fed to a fully connected neural network for the final classification. CNN depends on weight sharing, which performs well in favor of large images without the need to have a dedicated weight going across each pixel and the network's hidden layer. Weight sharing does not preserve the spatial locality of the learned features, which is practical for images, but not for time-frequency representation. This is related to the influence of the location of the detected feature at a specific frequency as a property to distinguish between sounds. The work of Abdel-Hamid et al. \citep{RN44} approached this problem by redesigning the convolutional filters to operate over bands. Another attempt was in \citep{RN340}, where they proposed using separate filters to convolve each of the time and frequency dimensions separately combined in the same model.

The Masked ConditionaL Neural Network (MCLNN) was introduced in \citep{RN465} with an analysis of the influence of the data split on model accuracy. In this work, we further evaluate the MCLNN performance on the music genre classification task.
\vskip -0.1in
\section{Conditional Neural Networks} 
\vskip 0in

The ConditionaL Neural Network (CLNN) \citep{RN465} is a discriminative model that extends from the generative Conditional Restricted Boltzmann Machine (CRBM) \citep{RN207} discussed earlier. The CLNN adapts the conditional previous visible to hidden links proposed in the CRBM, and it further extends the connections to the future frames as presented in the ICRBM \citep{RN257}. 

The CLNN is formed of a vector shaped hidden layer, similar to a conventional multi-layer perceptron, having \textit{e} dimensions. The input layer accepts a number of frames in a window of size \textit{d}, where the window's middle frame is conditioned on the past and future frames. The width of the window follows (\ref{eq:windowsize}) 
\vskip -0.25in
\begin{align}
\label{eq:windowsize}
d= 2n + 1 \qquad ,\,n \geq 1
\end{align}
\vskip -0.1in
where the 1 refers to the window's central frame and the \textit{n} frames refer to the neighboring frames to the middle one (2 is to account for the past and future directions). 
There are dense connections between each vector in the input window and the hidden layer. Accordingly, there are $2n+1$ weight matrices forming a tensor.
The weight tensor dimensions are [feature vector length \textit{l}, hidden layer width \textit{e}, window's depth \textit{d}]. Each vector of length \textit{l} in the input window of size \textit{d} has a corresponding dedicated weight matrix in the weight tensor. The new vectors generated from the vector-matrix multiplication between each feature vector and its corresponding weight matrix are summed together feature-wise before applying a nonlinear transformation. The activation of a hidden node is given in (\ref{eq:hidden_node_activation})
\vskip -0.25in
\begin{align}
\label{eq:hidden_node_activation}
y_{j,\; t} = f\left( b_{j} + \sum_{u=-n}^{n}\sum_{i=1}^{l} x_{i,\;u+t} \ W_{i,\;j,\;u}  \right)
\end{align}
\vskip -0.1in
where $y_{j,\;t}$ is the activation at node $j$ of the hidden layer for the window's middle frame at index \textit{t} of the segment. The segment, discussed later in detail, is a chunk of frames of a minimum size equal to the window. $f$  is the transfer function and $b_j$ is the bias at the $j^{th}$ node.  $x_{i,\;u+t}$ is the $i^{th}$ feature of the feature vector \textit{x}. \textit{u} refers to the index within the window and \textit{t} refers to the window's middle frame (having \textit{u}=0 in the window), which is at the same time the index of the middle frame in the input segment. $W_{i,\;j,\;u}$ is the weight between the $i^{th}$ feature of the vector at position \textit{u} in the window and the $j^{th}$ neuron in the hidden layer. \textit{u} is the index of a frame in the window and also the index of its corresponding weight matrix in the weight tensor. The hidden layer activation can be reformulated in a vector form in (\ref{eq:hidden_vector}).
\vskip -0.25in
\begin{align}
\label{eq:hidden_vector}
\hat{y}_{t}=f\left(\hat{b} + \sum_{u=-n}^{n} \hat{x}_{u+t}  \cdot  \hat{W}_u  \right) 
\end{align}
\vskip -0.1in

where the hidden layer activation vector $\hat{y}$ for the window's middle frame $x_{t}$ conditioned on the \textit{n} neighboring frames in either direction is given by the transfer function $f$, the bias vector $\hat{b}$ and the $vector-matrix$ multiplication between the feature vector $\hat{x}_{u}$  at index $u$ and its corresponding weight matrix $\hat{W}_u$ at the same index. The number of matrices in the weight tensor is equal to $2n+1$ matching the number of frames in the window, where each frame is processed by its dedicated matrix. The conditional distribution is formulated in $p(\hat{y_{t}}|\hat{x}_{-n+t},...,\hat{x}_{-1+t},\hat{x}_{t},\hat{x}_{1+t},...,\hat{x}_{n+t}) = \sigma(...) $, where $\sigma$ is a logistic function such as a Sigmoid or the output layer Softmax.

\begin{figure}
\vskip -0.25in
\begin{center}
\centerline{\includegraphics[width=7cm]{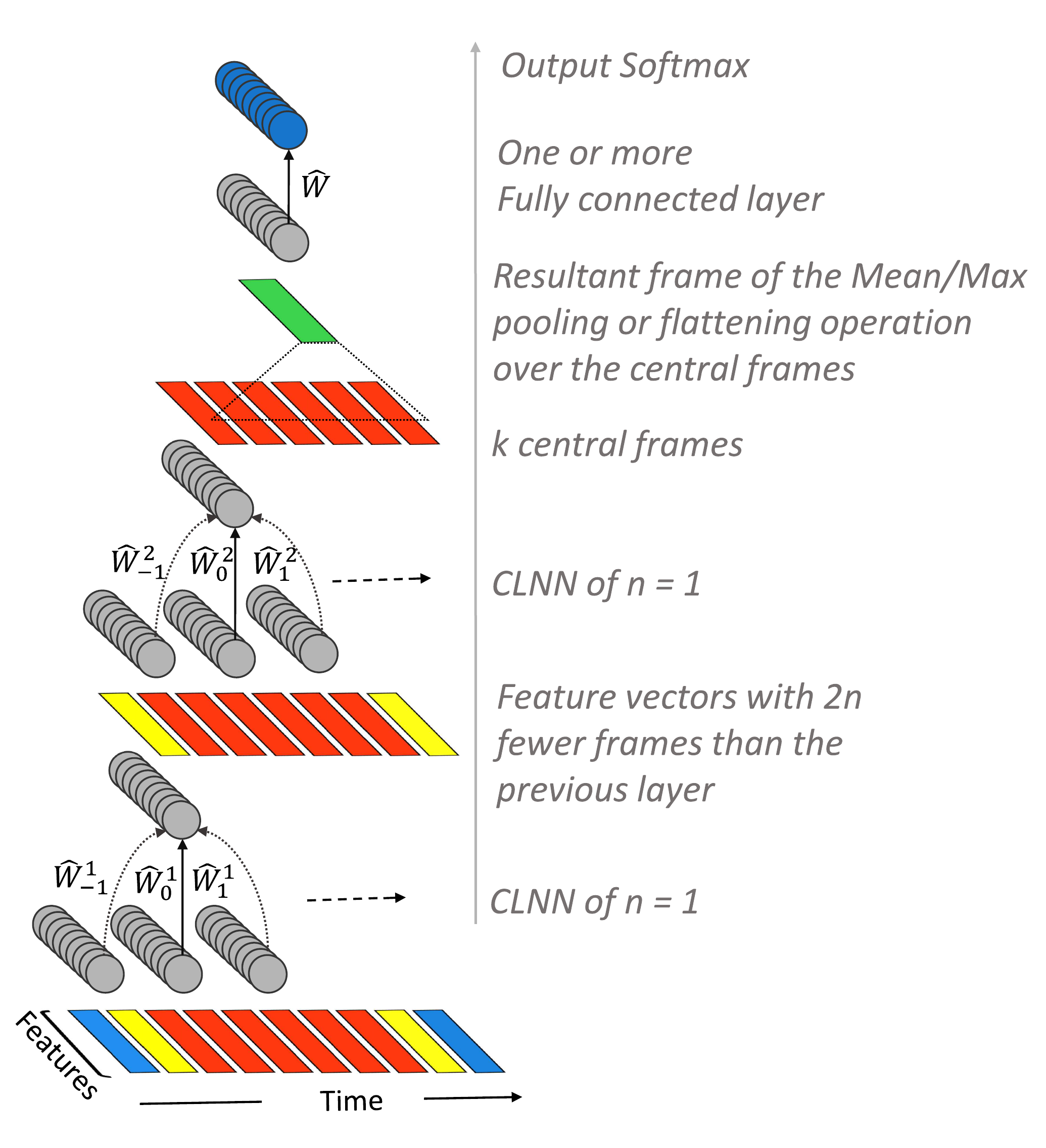}}
\vskip -0.2in
\caption{Two CLNN layers with $n=1$.}
\label{fig:clnn}
\end{center}
\vskip -0.5in
\end{figure}

Fig. \ref{fig:clnn} shows two CLNN layers of order $n=1$ followed by a global pooling layer \citep{RN332} that aggregates the features over $k$ extra frames before feeding them to a fully connected network for classification. Each CLNN layer consumes 2n frames generating a fewer number of frames. Accordingly, a CLNN is trained over segments of size following (\ref{eq:segments}) 
\vskip -0.25in
\begin{align}
\label{eq:segments}
q=(2n)m + k \qquad ,\,n,\,m\;\textrm{and}\;k \geq 1
\end{align}
\vskip -0.1in
where $q$ is the segment size, the order $n$ is for the number of frames in a single direction (the 2 is for the past and future frames ),  $m$ is the number of layers, and $k$ is for the extra frames to be pooled across beyond the CLNN layers. For example, at $n=4$, $m=3$ and $k=5$, a segment of size $(2\times4)\times3+5=29$ frames is presented at the input of the first CLNN layer. The second CLNN layer will receive $29-(2\times4)=21$ vectors at its input and consequently will generate $21-(2\times4)=13$ vectors as an output. Similarly, the third layer will generate $13-(2\times4)=5$ vectors, which undergo flattening or pooling to a single vector before the fully-connected layers. 

\section{Masked Conditional Neural Networks}
\vskip 0in
Spectrograms represent the energy at different frequency bins as the signal progresses through time. Despite the usefulness of such representations for signal analysis, they are susceptible to the frequency shifts, which could provide different spectral representations for very similar sounds. Frequency shift involves a smearing in the energy of a frequency bin across nearby bins due to uncontrolled factors affecting the signal propagation. Filterbanks tackle the frequency shifts in raw spectrograms. A filterbank is a group of filters used to subdivide the spectrograms into frequency bands allowing the new representation to be frequency shift-invariant. They are the principal operating component of Mel-scaled transformations such as the MFCC. The Masked ConditionaL Neural Networks (MCLNN) \citep{RN465} embed a filterbank-like behaviour within the network by enforcing a systematic sparseness over the network's links that follows a band-like pattern.

\begin{figure}[t]
\vskip 0in
\begin{center}
\centerline{\includegraphics[width=\textwidth]{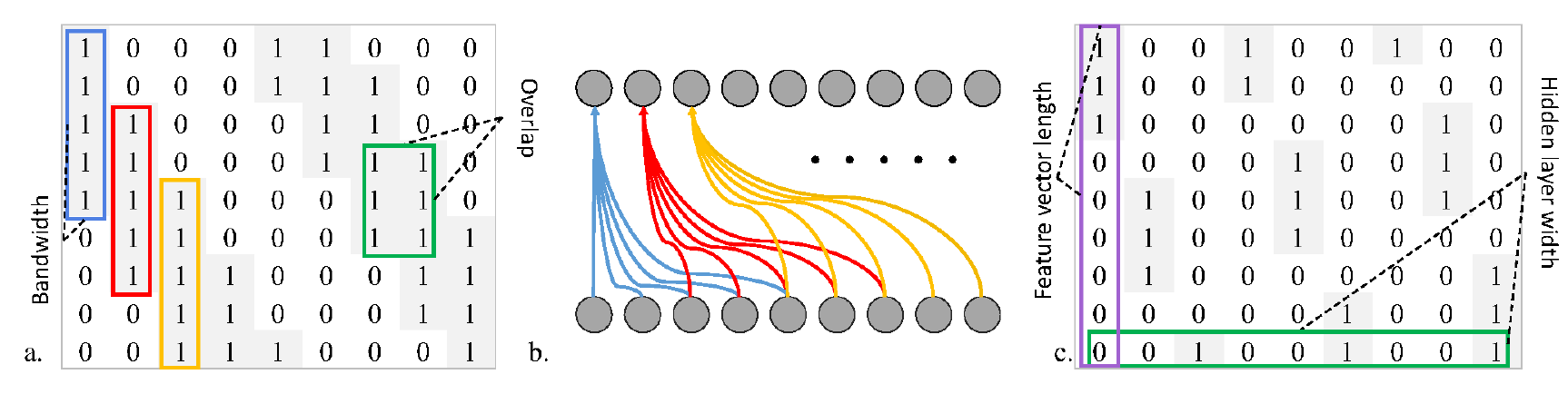}}
\vskip -0.2in
\caption[9pt]{Masking patterns. a) $Bandwidth = 5$ and $Overlap = 3$, b) the active links following the masking pattern in a. c) $Bandwidth = 3$ and $Overlap = -1$ }
\label{fig:mask}
\end{center}
\vskip -0.7in
\end{figure}

The mask design is controlled by two tunable hyper-parameters: the Bandwidth and the Overlap. Fig. \ref{fig:mask}.a. shows a masking pattern with a Bandwidth of 5 and an Overlap of 3. The Bandwidth values refer to the successive 1's in a column, and the Overlap refers to the superposition of the patterns between one column and another. Fig. \ref{fig:mask}.b. depicts the active connections following the mask in Fig. \ref{fig:mask}.a. Each neuron in the hidden layer of Fig. \ref{fig:mask}.b.  has a focused spatial region of the feature vector to observe. Fig. \ref{fig:mask}.c. shows a mask with a negative overlap depicting the non-overlapping distance between two columns. The linear indexing of the binary values of a mask is formulated in (\ref{eq:linearindex}) 
\vskip -0.25in
\begin{align}
\label{eq:linearindex}
lx=a + (g-1)(l+(bw-ov))
\end{align}
\vskip -0.1in

where the linear index $lx$ is given by the bandwidth $bw$, the overlap $ov$ and the feature vector length $l$. $a$ takes the values in $[\;0, \; bw-1\;]$ and $g$ is in the interval $[  \;1, \lceil (l \times e)/(l+(bw-ov))\rceil\;  ]$. The mask plays another role of exploring a range of feature combinations analogous handcrafting the optimum feature combinations. This operation is applied in the MCLNN for several feature combinations concurrently as shown in Fig. \ref{fig:mask}.c., where the 2nd set of three columns holds a shifted version of the 1st three columns and similarly for the 3rd set. In a closer analysis, each hidden node (mapped to a column in the mask) will have a different input to observe. For example, the input at the 1st node is the first three features of the feature vector, the 4th node's input is the first two features, and the 7th node is the first feature. The masking is applied through an element-wise multiplication following (\ref{eq:mask}).
\vskip -0.2in
\begin{align}
\label{eq:mask}
\hat{Z}_u =\hat{W}_u \circ \hat{M}
\end{align}
\vskip -0.1in
where $\hat{W}_u$ is the original weight matrix at index $u$, $\hat{M}$ is the masking pattern and $\hat{Z}_u$ is the masked weight matrix to replace the original one in (\ref{eq:hidden_vector}).

\vskip -0.3in
\begin{figure}
\vskip 0in
\begin{center}
\centerline{\includegraphics[width=5.5cm]{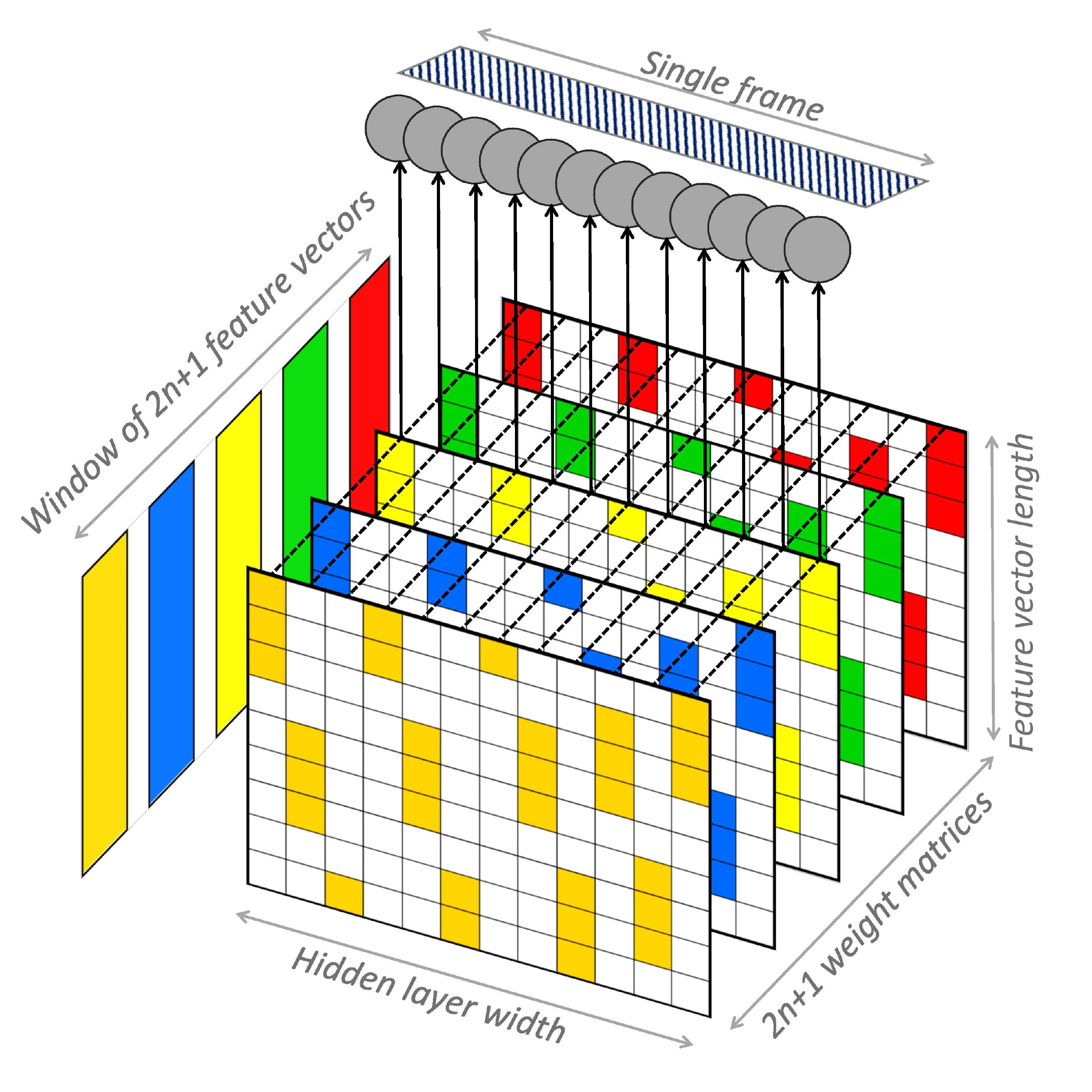}}
\vskip -0.1in
\caption{Single MCLNN step.}
\label{fig:mclnn}
\end{center}
\vskip -0.5in
\end{figure} 

Fig. \ref{fig:mclnn} shows a single MCLNN step, where a window of frames of size $2n+1$ is processed with a matching count of matrices. Each frame in the window has a corresponding matrix to process. The vector-matrix multiplication generates \textit{d} new vectors, which are summed feature-wise before applying the nonlinearity by a transfer function. The output of a single step over the window is a resultant single frame. The highlighted cells in each matrix depict the active links enforced through the mask.
\vskip -0.5in
\section{Experiments}
\vskip 0in
We evaluated the performance of the MCLNN using the Ballroom \citep{RN425} and the Homburg \citep{RN437} datasets widely adapted for Music Information Retrieval tasks including genre classification. 

The \textit{Ballroom} dataset is composed of 698 music clips of 30 seconds each, unevenly partitioned across 8 music genres: Cha Cha (CC), Jive (Ji), Quickstep (Qs), Rumba (Ru), Samba (Sa), Tango (Ta), Viennese Waltz (VW) and Slow Waltz (SW). 

The \textit{Homburg} dataset contains 1886 music clips of 10 seconds each, distributed across 9 classes: Alternative (Al), Blues (Bl), Electronic (El), FolkCountry (FC), FunkSoulRnb (FS), Jazz (Ja), Pop (Po), RapHiphop (RH) and Rock (Ro).

All files for both datasets were transformed to a logarithmic mel-scaled spectrogram of 256 bin using an FFT of 2048 and 1024 hop size. Segments were extracted following (\ref{eq:segments}) and the z-score parameters of the training data were used to standardize the testing and validation sets. Experiments were carried out using a 10-folds cross-validation with the mean accuracy across the folds reported. The hyper-parameters used for the MCLNN are listed in Table \ref{table:mirmodelparam}. 

\begin{table}
       \centering
       \caption{MCLNN hyper-parameters for the Ballroom and the Homburg}
  	   \label{table:mirmodelparam}
  	   \vskip -0.2cm
       \begin{tabular}{lccccc}
\hline
\addlinespace[0.1cm]
\parbox[t]{1cm}{\multirow{2}{*}{\centering Layer}}  & \parbox[t]{1cm}{\centering Hidden \\ Nodes} &  \parbox[t]{1.5cm}{\centering Mask \\ Bandwidth}   & \parbox[t]{1.2cm}{\centering Mask \\ Overlap} & \parbox[t]{1.4cm}{\centering Order $n$ \\ \textit{(Ballroom)}} & \parbox[t]{1.4cm}{\centering Order $n$ \\ \textit{(Homburg)}}  \\[4ex] 
\addlinespace[-0.1cm]
\hline
\addlinespace[0.1cm]
1    & 220 & 40 & -10	& 15	& 5\\
2    & 200 & 10 & 3		& 15	& 5\\
\hline
		\end{tabular}
\vskip -0.2in
\end{table}

The two MCLNN layers are followed by a global single dimension pooling layer to pool feature-wise over the $k$ extra frames. The global pooling emulates the aggregation over a musical texture window, which was studied by Bergstra et al. \citep{RN295}. We used $k=11$ and  $k=2$ for the Ballroom and the Homburg, respectively. Two densely connected layers of 50 and 10 nodes followed the global pooling layer, before the final Softmax. The model was trained using ADAM \citep{RN308} to minimize the categorical cross-entropy between the predicted vector and the target label. Dropout \citep{RN311}  was used as a regularizer. The final decision of the clip's category is decided using probability voting across the frames of the clip.
\vskip -0.2in
\begin{table}
    \begin{minipage}{0.5\linewidth}
     \caption{Reported accuracies on Ballroom }
  \label{table:ballroom}
      \centering
      \begin{tabular}{lc}
        \hline\rule{0pt}{12pt}
        \parbox[t]{3.0cm}{ Classifier and Features} & Ac.\%\\
        \midrule
SVM + 28 feature,Tempo \citep{RN335} &96.13\\
KNN + Modulation Scale Spec. \citep{RN416}&93.12\\
Manhattan Dist. + Block-Level feat. \citep{RN333}&92.44\\
\textbf{MCLNN + Mel$-$Spec.(this work)}&\textbf{90.40}\\
SVM + Rhyth.,Hist.,Stat.,Onset,etc\citep{RN341}&90.40\\
KNN + 15 MFCC-like desc.,Tempo\citep{RN415}&90.10\\
KNN + Rhythm and Timbre\citep{RN343}&89.20\\
SVM + 28 features without Tempo \citep{RN335}& 88.00\\
CNN+ Mel-Scaled Spectrogram\citep{RN340}&87.68\\
SVM + Rhyth.,Hist.,Statist. \citep{RN339}&84.20\\
KNN + Tempo \citep{RN415}&82.30\\
       \hline\rule{0pt}{12pt}     
      \end{tabular}
    \end{minipage}
    \begin{minipage}{0.47\linewidth}
		\caption{Reported accuracies on Homburg }
		\label{table:homburg}
		\centering
		\begin{tabular}{lc}
		\hline\rule{0pt}{12pt} 
        \parbox[t]{3cm}{Classifier and Features} & Ac.\%\\
        \midrule
        JSLRR+Cortrical Representations \citep{RN434}&63.46\\
        LRSM+Cort.,MFCC,Chro.\citep{RN440}&62.40\\
\textbf{MCLNN + Mel$-$Spec. (this work)}&\textbf{61.45}\\
KNN+LFP,VDSP,CP,SCP\citep{RN441}&61.20\\
SVM+ESA-MFCC\citep{RN442}&57.81\\
KNN + Rhythm and Timbre\citep{RN343}&57.00\\
KNN+mcRBM,PCA,MVG-MFCC\citep{RN443}&55.30\\
SVM+Marsyas features(\citep{RN294})\citep{RN444}&55.00\\
KNN+Multiple features\citep{RN437}&53.23\\
SVN + Novelty Functions \citep{RN280}&51.10\\
KNN+ mcRBM,PCA,Mel-Spec.\citep{RN433}&45.50\\
        \hline\rule{0pt}{12pt}
      \end{tabular}
    \end{minipage}
\vskip -0.45in
\end{table}

As listed in Table \ref{table:ballroom} and Table \ref{table:homburg}, MCLNN achieved an accuracy of 90.4\% and 61.45\% on the Ballroom and the Homburg, respectively, which surpasses several neural network based architectures in addition to hand-crafted attempts on both datasets. MCLNN achieved the mentioned accuracies without a special design to exploit musical perceptual properties compared to other attempts. In the work of Peeters \citep{RN335}, he achieved $96.13\%$ on the Ballroom using the Tempo annotations released with the dataset. Peeters reapplied his proposed handcrafted features without Tempo data, and the accuracy was $88\%$, which shows the influence of the tempo annotations. In a similar type of analysis, Gouyon et al.\citep{RN415} used the Tempo annotations as a baseline to benchmark their proposed handcrafted features, where the Tempo annotations alone achieved $82.3\%$ and their proposed features with the Tempo achieved $90.1\%$. The work of Marchand et al. \citep{RN416} achieved $93.12\%$ using multiple processing stages including on-set energy calculation, autocorrelation, modulation scale spectra and dimensionality reduction to exploit rhythmic pattern in a music clip. Seyerlehner et al. \citep{RN333} achieved $92.44\%$ using several features extracted from blocks of the spectrogram. A neural network based attempt in the work of Pons et al. \citep{RN340} achieved $87.68\%$ using a shallow CNN architecture with pre-trained filters convolving the time and spectral dimensions separately in the same model.  
Handcrafted features for the Homburg dataset has been explored as well. The work of Panagakis et al. \citep{RN434} achieved $63.46\%$ using the auditory cortical representations in combination with their introduced classifier. Their work reports the accuracy achieved on the Ballroom dataset using the same features (cortical representations) and the classifier used for the Homburg, where they achieved $81.93\%$ on the Ballroom dataset. The work in \citep{RN440} achieved $62.4\%$ on the Homburg dataset using auditory cortical representations, MFCC and Chroma as features. A neural network based attempt on the Homburg dataset in the work of Schluter et al. \citep{RN433} achieved $45.5\%$ using mcRBM \citep{RN470}, a variant of the RBM, applied on a mel-spectrogram.
 
\begin{figure}
\begin{minipage}{0.48\textwidth}
\vskip -0.1in
\begin{center}
\centerline{\includegraphics[width=6cm]{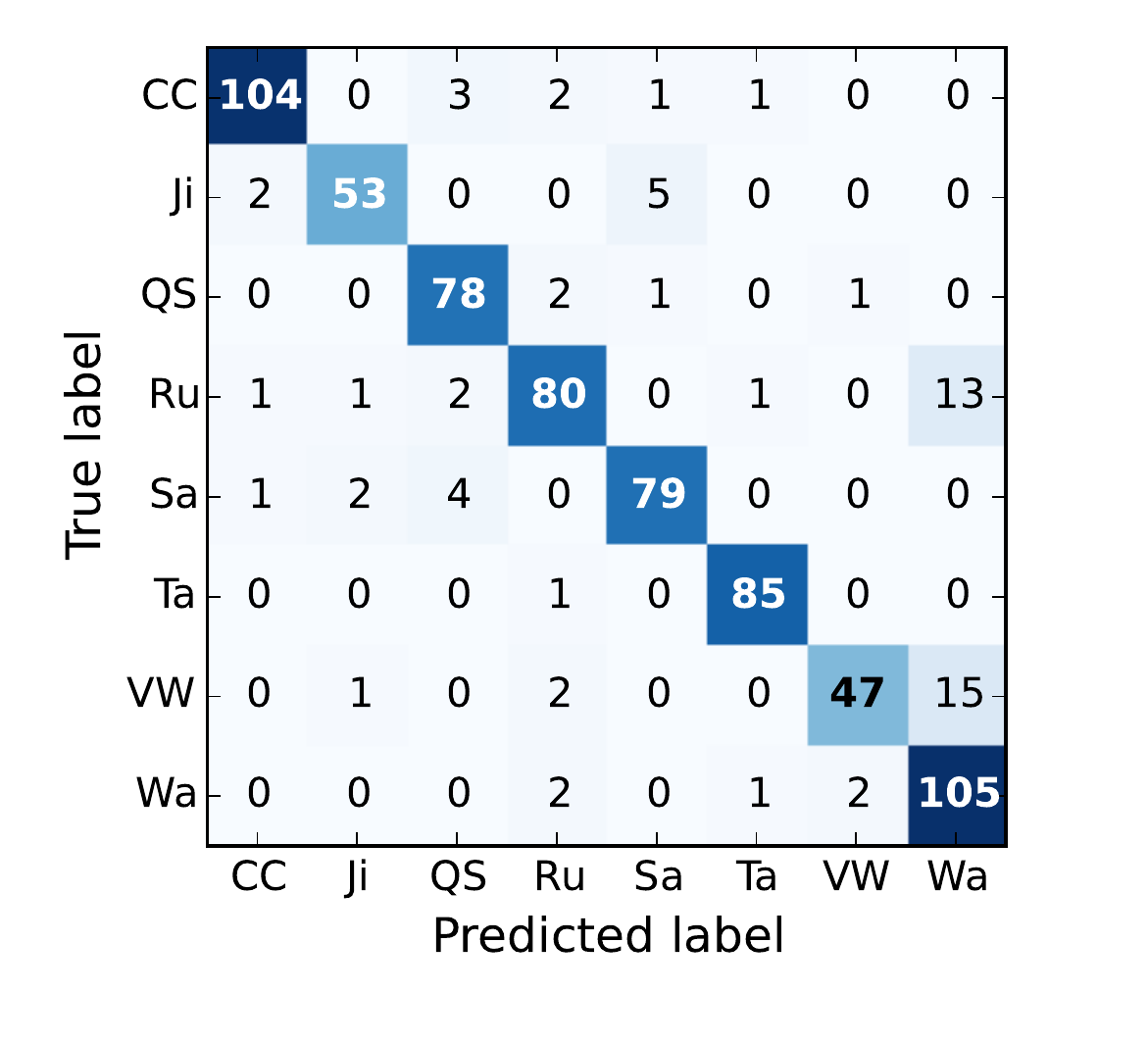}}
\vskip -0.2in
 \caption{\footnotesize{Ballroom confusion using the MCLNN.}}
\label{fig:confballroom}
\end{center}
\end{minipage}\hfill
\begin{minipage}{0.48\textwidth}
\begin{center}
\centerline{\includegraphics[width=5.5cm]{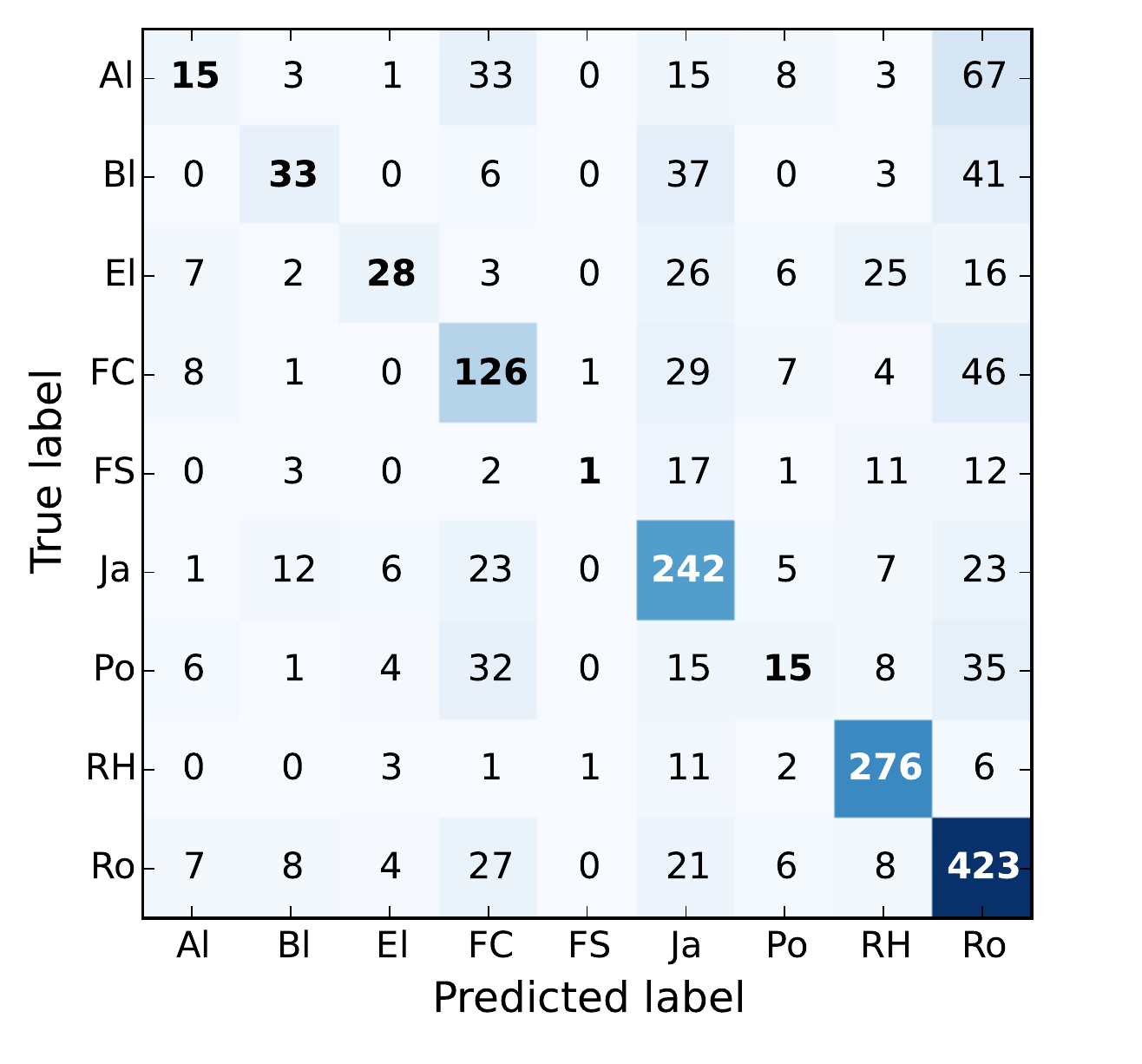}}
\vskip -0.2in
\caption{\footnotesize{Homburg confusion using the MCLNN.}}
\label{fig:confhomburg}
\end{center}
\end{minipage}\hfill
\vskip -0.4in
\end{figure}

Fig.~\ref{fig:confballroom} and Fig. \ref{fig:confhomburg} show the confusion matrix for Ballroom and the Homburg datasets, respectively. High confusion is noticed for the Rumba and the Waltz genres with the Slow Waltz, which overlap with the findings in \citep{RN280}. For the Homburg dataset, less confusion is noticed with the availability of more samples in the genre category.
\vskip -0.2in

\section{Conclusions and Future work}
\vskip 0in
In this work, we have explored the applicability of the ConditionaL Neural Network (CLNN) and the Masked ConditionaL Neural Network (MCLNN) on the music genre classification task. The CLNN preserves the inter-frames relation of a temporal signal and the spatial locality of the features. The MCLNN extends the CLNN by enforcing a systematic sparseness over the network's links following a band-like pattern, which mimics a filterbank. The filterbank-like pattern induces the network to learn in frequency bands. The mask also automates the exploration of several feature combinations concurrently, which is usually a manual process of handcrafting the optimum feature combinations. The MCLNN has achieved competitive accuracies on the Ballroom and the Homburg music datasets compared to several handcrafted attempts, in addition to state-of-the-art Convolutional Neural Networks. The MCLNN has achieved these accuracies without depending on any musical perceptual properties used in several hand-crafted attempts, which allow the MCLNN to generalize to other types of multi-dimensional temporal signals. Future work, we will consider using deeper MCLNN architectures with more optimization to the masking patterns used, in addition to using different orders across the layers. We will also explore applying the MCLNN to multi-dimensional temporal representations other than spectrograms.

\subsubsection*{Acknowledgments.} This work is funded by the European Union's Seventh Framework Programme for research, technological development and demonstration under grant agreement no. 608014 (CAPACITIE).
\bibliographystyle{splncs03}

\bibliography{referencefile}

\end{document}